\documentclass{article}

    \usepackage[final, nonatbib]{tackling_climate_workshop_style}

\usepackage[utf8]{inputenc} 
\usepackage[T1]{fontenc}    
\usepackage[breaklinks=true,colorlinks,bookmarks=false]{hyperref}
\usepackage{url}            
\usepackage{booktabs}       
\usepackage{amsfonts}       
\usepackage{nicefrac}       
\usepackage{microtype}      
\usepackage{subcaption}
\usepackage{booktabs} 
\usepackage{amsmath}
\usepackage{multirow}
\usepackage{graphicx}
\usepackage[style=numeric,sorting=none]{biblatex}
\title{Typhoon Intensity Prediction with Vision Transformer}

\author{
  Huanxin Chen \\
  South China University of Technology \\
  Guangzhou, China\\
   \And
  Pengshuai Yin \\
  Guangdong Laboratory of Artificial \\
  Intelligence and Digital Economy \\
  Shenzhen, China \\
   \And
  Huichou Huang \\
  City University of Hong Kong \\
  Hong Kong, China \\
   \And
  Qingyao Wu \\
  South China University of Technology \\
  Guangzhou, China \\
   \And
  Ruirui Liu \\
  Brunel Univeristy London \\
  London, United Kingdom \\
   \And
  Xiatian Zhu\thanks{Xiatian Zhu (xiatian.zhu@surrey.ac.uk) is the corresponding author.\\  
  \indent \, {\bf Acknowledgement}: This work is supported by China Postdoctoral Science Foundation (2022M721182)} \\
  CVSSP, University of Surrey \\
  Guildford, United Kingdom \\
}

\begin{document}
\maketitle

\begin{abstract}

Predicting typhoon intensity accurately across space and time is crucial for issuing timely disaster warnings and facilitating emergency response. This has vast potential for minimizing life losses and property damages as well as reducing economic and environmental impacts. Leveraging satellite imagery for scenario analysis is effective but also introduces additional challenges due to the complex relations among clouds and the highly dynamic context.
Existing deep learning methods in this domain rely on convolutional neural networks (CNNs), which suffer from limited per-layer receptive fields. This limitation hinders their ability to capture long-range dependencies and global contextual knowledge during inference.
In response, we introduce a novel approach, namely ``Typhoon Intensity Transformer'' ({\bf Tint}), which leverages self-attention mechanisms with global receptive fields per layer. Tint adopts a sequence-to-sequence feature representation learning perspective. It begins by cutting a given satellite image into a sequence of patches and recursively employs self-attention operations to extract both local and global contextual relations between all patch pairs simultaneously, thereby enhancing per-patch feature representation learning.
Extensive experiments on a publicly available typhoon benchmark validate the efficacy of Tint in comparison with both state-of-the-art deep learning and conventional meteorological methods.
Our code is available at \url{https://github.com/chen-huanxin/Tint}.

\end{abstract}

\section{Introduction}


In recent decades, global warming has led to both the intensification and expansion of typhoons \cite{sun2017impact}. Typhoons are severe or even extreme weather systems that originate from warm tropical oceans and gradually build up their own power when approaching nearby lands. Upon making landfall, they pose significant threats to lives and properties in their vicinity \cite{rappaport2000loss,mendelsohn2012impact}. Different typhoon intensities correspond to varying levels of economic losses and environmental devastation \cite{zhai2014dependence}. The intensity of a typhoon is closely associated with the maximum sustained surface wind speed near its center, making it a key factor in disaster management.

Typhoon satellite imagery plays a pivotal role in predicting typhoon intensity due to its rich and timely real-time data, enabling us to dynamically monitor the typhoon's structure, cloud patterns, and environmental conditions \cite{velden2006dvorak}. Ample studies focus on predictive regressions using the information, features, and parameters extracted from these remote sensing images \cite{velden1998development,fetanat2013objective}.

In recent years, to overcome the limitations of regression-based methods, researchers have resorted to deep learning methods \cite{lecun2015deep} for estimating typhoon intensity using satellite imagery \cite{miller2017using}.
One of the most prevailing methods in this field relies heavily on Convolutional Neural Networks (CNNs) \cite{he2016deep}, which are renowned for their proficiency in capturing local image features and intricate image structures. However, the over-concentration on local information may deteriorate the model's overall performance as it neglects the global context that is essential for improving predictive accuracy.

In this study, we address the above critical limitations by proposing a novel method called the {\em Typhoon Intensity Transformer} ({\bf Tint}). The Tint employs self-attention mechanisms with expansive global fields in each layer. This is achieved by partitioning input images into fixed-size patches and transforming them into one-dimensional feature vector sequences. Self-attention mechanisms are then used for establishing the connections among these patches so as to extract comprehensive global features and contextual information spanning the entire image. Moreover, The Tint further refines and consolidates image features through the incorporation of multiple Transformer layers, through which these features are forwarded to an output layer for typhoon intensity predictions. The superior performance of the Tint can be attributed to its capacity in treating images analogously to text data from a sequential perspective, i.e., it is able to extract extensive global relations and contextual cues beyond the local features captured by the CNNs.

\section{Methodology}

\begin{figure}
    \centering
    \includegraphics[width=135mm]{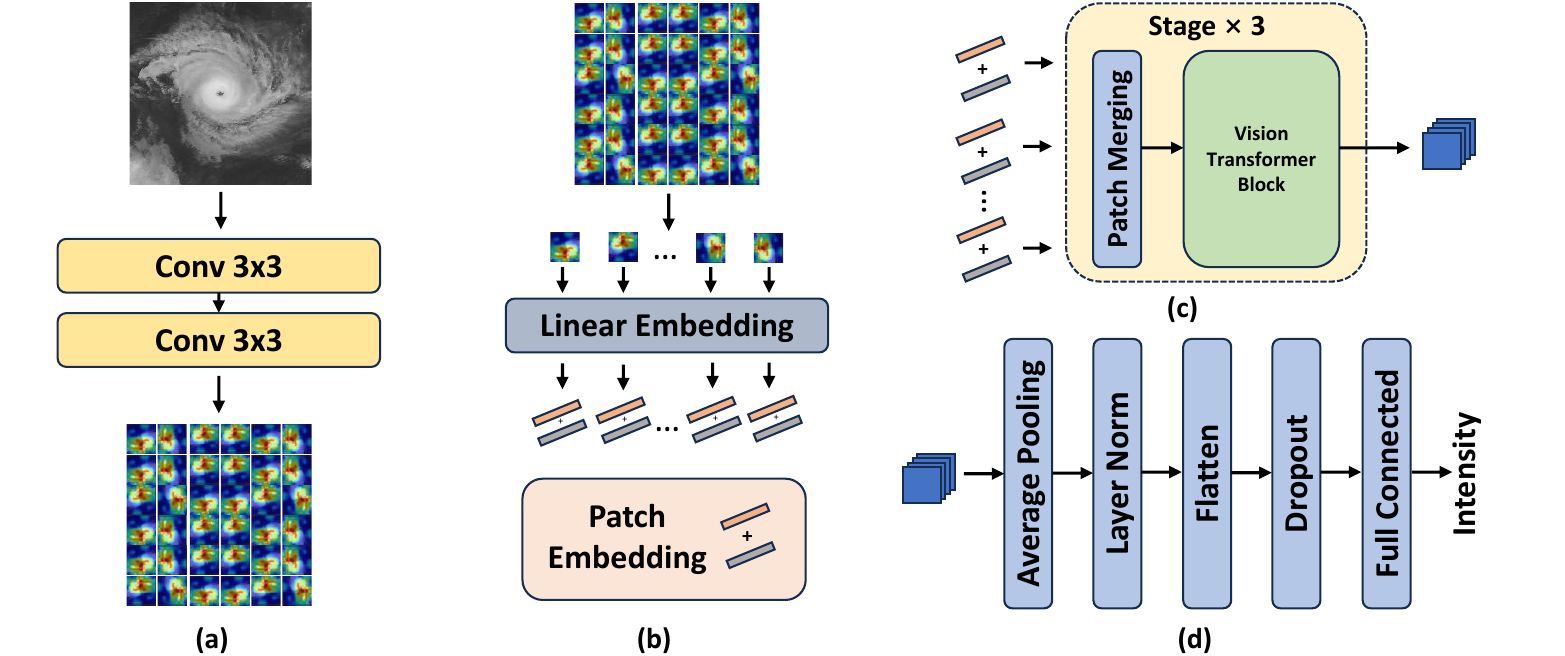}
    \caption{Overview of our proposed {\em Typhoon Intensity Transformer} (Tint) method. 
    Our approach begins with a sensor image, which undergoes the following key steps:
(a) An input image is initially divided into a grid of patches.
(b) We then perform patch embedding and integrate positional embeddings, resulting in a sequence of feature embeddings, which serve as our initial image representations.
(c) Subsequently, we employ a lightweight vision Transformer to process this sequence of feature embeddings. This step enables both local and global information processing, yielding a comprehensive representation for each patch.
(d) Finally, we apply average pooling to aggregate the patch-level feature representations, creating an image-level representation. This is followed by a fully connected layer, which is responsible for predicting the typhoon intensity.
    }
    \label{fig:pipeline}
\end{figure}

The Tint adopts the same input-output configuration as Vision Transformers\cite{tiny_vit,liu2021swin,graham2021levit} with a key difference that it outputs a continuous integer as the typhoon intensity estimation. Overview of our Tint is given in Figure~\ref{fig:pipeline}. 

\paragraph{Image to sequence} 
Our method takes a 1D sequence of feature embeddings $Z \in \mathbb{R}^{L\times C}$ as inputs with $L$ as the preset sequence length and $C$ as the hidden channel size. 
This is implemented as follows:
(1) We first cut a given input image into a grid of patches;
(2) Then an embedding block is employed to encode each individual patch
into a feature embedding;
(3) Positional embedding is further added on top of each content embedding.

\paragraph{Model architecture} 
To strike a balance between computational efficiency and model performance, we choose to implement the Tiny-ViT architecture \cite{tiny_vit}. Our model comprises four stages with progressively decreasing resolutions. The patch embedding block consists of two convolutional layers using a kernel size of 3, a stride of 2, and a padding size of 1. In the initial stage, we employ lightweight and efficient MBConvs \cite{howard2019searching} along with downsampling blocks. This choice is motivated by the efficiency of early convolutional layers in learning low-level features \cite{xiao2021early}.

The subsequent three stages are constructed using Transformer blocks that employ window attention mechanisms to reduce computational overhead. We incorporate attention bias \cite{graham2021levit} and apply a separable convolution of 3$\times$3 depth between the attention and MLP components to capture local information \cite{wu2021rethinking}, emphasizing the intrinsic ability to learn global context information. The residual connections \cite{he2016deep} are used in each block of the first stage also within the attention and MLP blocks. For activation functions, we employ GELU \cite{hendrycks2016gaussian}. Furthermore, BatchNorm \cite{ioffe2015batch} is applied to the normalization layers of the convolutional layers, and LayerNorm \cite{ba2016layer} is used for the linear layers.

The architectural selection aims to balance computational efficiency and model performance. Formally, given the embedding sequence $Z$ as an input,
the Transformer with $M$ layers of multi-head
self-attention (MSA) and Multilayer Perceptron (MLP) blocks is employed to learn feature representations. 
At each layer $l$, the input to self-attention is a triplet
(\texttt{query}, \texttt{key}, \texttt{value})
computed from the input $Z^{l-1} \in \mathbb{R}^{L\times C}$ as ${Z^{l-1}} \textbf{W}_q$, ${Z^{l-1}} \textbf{W}_k$, and ${Z^{l-1}} \textbf{W}_v$, respectively,
where $\textbf{W}_q$/$\textbf{W}_k$/$\textbf{W}_v  \in \mathbb{R}^{C\times d}$
are the parameters of the learnable linear projections,
and $d$ is the dimension of (\texttt{query}, \texttt{key}, \texttt{value}).
Self-attention (SA) is then formulated as:
\begin{equation}
    SA(Z^{l-1}) = Z^{l-1} + 
    \operatorname{softmax}(\frac{Z^{l-1} \textbf{W}_q (Z^{l-1} \textbf{W}_k)^\top}{\sqrt{d}}) (Z^{l-1} \textbf{W}_v).
    \label{eq:attn}
\end{equation}
The Multi-Head Self-Attention (MSA) design including $m$ independent SA operations is deployed.
It projects a concatenated output in the form of
$MSA(Z^{l-1}) = [SA_1(Z^{l-1});~SA_2(Z^{l-1}); ~\cdots;~SA_m(Z^{l-1})]\textbf{W}_O$, where $\textbf{W}_O \in \mathbb{R}^{md \times C} $ and $d$ is set to $C/m$.
The output of the MSA is then transformed 
by an Multi-Layer Perceptron (MLP) block with a residual skip as the layer output $Z^l = MSA(Z^{l-1}) + MLP(MSA(Z^{l-1})) \in \mathbb{R}^{L \times C}$.
The layer normalization applied prior to the MSA and MLP blocks is excluded here for simplicity.

\paragraph{Head design}

The final layer's feature output is represented as $Z^m \in \mathbb{R}^{L \times C}$. We initially perform an average pooling operation on $Z^m$, generating $ {\bf v} = avg(Z^m)$.
Subsequently, a fully connected layer uses ${\bf v}$ to derive the final prediction $y = {\bf v} {\bf w} + b$ with ${\bf w}$ and $b$ as trainable parameters.

\section{Experiments}
\subsection{Experiment Settings}

{\bf Dataset} 
Our experiments are conducted using the Tropical Cyclone for Image-to-Intensity Regression (TCIR) dataset \cite{chen2018rotation}. This dataset serves as an important open benchmark for evaluating typhoon intensity estimation models with a fair assessment setting. See the Supplementary Dataset for details.

{\bf Performance metric}
Given that typhoon intensity estimation is formulated as a regression problem \cite{chen2018rotation}, we evaluate the model's performance using the Root Mean Squared Error (RMSE). 

{\bf Training strategy}
Details of our implementation can be found in the \texttt{Supplementary Material}.


{\bf Competitors}
We conduct a comparative analysis of our Tint model with conventional meteorological models and recent deep-learning methods (See more details of these competitors in the \texttt{Supplementary Material}).

\begin{table}[htbp]
  \caption{Compare with other models. IR: Infrared; PMW: Passive Microwave; WV: Water Vapor. The unit of RMSE is knots}
  \begin{subtable}{0.5\textwidth} 
  \caption{Performance evaluation on TCIR validation dataset.}
    \label{sample-table2}
    \centering
    \begin{tabular}{lll}
      \toprule
    {\centering Model} & {\centering Modality} & {\centering RMSE}\\
    \midrule
    ADT \cite{velden1998development} & IR & 11.79\\
    AMSU \cite{kidder2000satellite} & IR & 14.10\\
    SATCON \cite{velden2014update} & IR+PMW &\bf 9.21\\
    KT \cite{kossin2007globally} & IR+PMW & 13.20\\
    FASI \cite{fetanat2013objective} & IR & 12.70\\
    Improved DAV-T \cite{ritchie2014satellite} & IR & 12.69\\
    TI index \cite{liu2015satellite} & IR & 9.34\\
    \midrule
    MLRM \cite{zhao2016multiple} & IR & 12.01\\
    TDCNN \cite{miller2017using} & IR & 10.00\\
    Deep CNN \cite{pradhan2017tropical} & IR & 10.18\\
    CNN-TC \cite{chen2018rotation} & IR & 10.59\\
    GAN-CNN \cite{chen2021real} & IR & 10.45\\
    ResNet32 \cite{he2016deep} & IR & 11.63\\
    \bf Tint & IR & 9.63\\
    \bf Tint & IR+PMW & \bf 9.54\\
    \bottomrule
    \end{tabular}
  \end{subtable}
  \begin{subtable}{0.5\textwidth} 
  \caption{Performance evaluation on TCIR testing dataset.}
    \label{sample-table3}
    \centering
    \begin{tabular}{lll}
      \toprule
        {\centering Model} & {\centering Modality} & {\centering RMSE}\\
        \midrule
        ADT \cite{velden1998development} & IR & 12.19\\
        SATCON \cite{velden2014update} & IR+PMW & 9.21\\
        \midrule
        CNN-TC \cite{chen2018rotation} & IR+PMW & 10.13\\
        GAN-CNN \cite{chen2021real} & IR+WV & 10.45\\
        ResNet32 \cite{he2016deep} & IR & 10.42\\
        ResNet32 \cite{he2016deep} & IR+WV & 10.39\\
        ResNet32 \cite{he2016deep} & IR+PMW & 10.32\\
        \bf Tint & IR & 9.35 \\
        \bf Tint & IR+WV & 9.33 \\
        \bf Tint & IR+PMW & \bf 9.00 \\
        \bottomrule
    \end{tabular}
  \end{subtable}
\end{table}

\subsection{Evaluation results}

We present the results from typhoon intensity estimation using various methods in Table~\ref{sample-table2} (validation set) and Table~\ref{sample-table3} (testing set). Our analysis starts with a comprehensive examination of the validation set, offering the following insights:

{\bf Meteorological Modeling}: Surprisingly, simple linear regressions show considerable performance, surpassing more complex retrieval-based methods (e.g., KT \cite{kossin2007globally} and FASI \cite{fetanat2013objective}), as well as brightness temperature gradient-based approaches (e.g., Improved DAV-T \cite{ritchie2014satellite}). Notably, the TI index as one of the latter \cite{liu2015satellite} provides valuable incremental information but is slightly outperformed by the composite approach that leverages multiple models and meteorological expertise \cite{velden2014update}.

{\bf Deep Learning}: 
Most of the deep learning methods exhibit competitive performance or even surpass their conventional counterparts, highlighting the promising potential of the data-driven methods. In particular, ResNet \cite{he2016deep}, as one of the most widely used CNN architectures, shows notable performance comparable to traditional linear regressions. CNN models with specific task-oriented designs, such as regularization \cite{pradhan2017tropical} and multi-modality exploitation \cite{chen2018rotation}, exhibit improved performance. Still, these developments fail to overcome the limitations imposed by restricted receptive fields. While it is clear that our Tint model characterized by per-layer global receptive fields outperforms significantly. This supportive evidence validates our model assumptions and structure design.

Our findings remain consistent and qualitatively unchanged on the test set, confirming the stability and generality of our proposed model. In addition, the empirical results suggest that incorporating more sensor data tends to improve the predictive performance. 
In the test set, our model achieves state-of-the-art after adding WV information, and more so with PMW information.

{\bf Qualitative evaluation}:
We also provide attention visualization for comparing ResNet32 with our Tint model (refer to Figure~\ref{fig2:env} in the \texttt{Supplementary Material} for details). The visualization clearly indicates that Tint captures a broader spatial context and empowers more comprehensive feature representation learning that results in significant improvement in prediction accuracy.


\section{Conclusion}

In summary, our ``Typhoon intensity Transformer'' ({\bf Tint}) model significantly improves the predictive accuracy of typhoon intensity, which is of great value for real-world disaster management. The innovative use of self-attention mechanisms is attributable to the superior performance of the Tint, as they succeed in featuring global receptive fields per layer, thereby substantially enhancing its ability in capturing the long-range dependencies in sensor observations. This also makes it adaptive across diverse domains with dynamic high-dimensional data. The results from the experiments conducted on a benchmark dataset demonstrate Tint's superiority over both deep learning and meteorological methods and highlight Tint's potential in innovating disaster management in response to meteorological disasters such as typhoons.

\newpage
\small
\printbibliography

@article{velden1998development,
  title={Development of an objective scheme to estimate tropical cyclone intensity from digital geostationary satellite infrared imagery},
  author={Velden, Christopher S and Olander, Timothy L and Zehr, Raymond M},
  journal={Weather and Forecasting},
  volume={13},
  number={1},
  pages={172--186},
  year={1998}
}

@article{velden2014update,
  title={Update on the SATellite CONsensus (SATCON) algorithm for estimating TC intensity},
  author={Velden, CS and Herndon, D},
  journal={Poster session I. San Diego},
  year={2014}
}

@inproceedings{chen2018rotation,
  title={Rotation-blended CNNs on a new open dataset for tropical cyclone image-to-intensity regression},
  author={Chen, Boyo and Chen, Buo-Fu and Lin, Hsuan-Tien},
  booktitle={Proceedings of the 24th ACM SIGKDD International Conference on Knowledge Discovery \& Data Mining},
  pages={90--99},
  year={2018}
}

@article{kidder2000satellite,
  title={Satellite analysis of tropical cyclones using the Advanced Microwave Sounding Unit (AMSU)},
  author={Kidder, Stanley Q and Goldberg, Mitchell D and Zehr, Raymond M and DeMaria, Mark and Purdom, James FW and Velden, Christopher S and Grody, Norman C and Kusselson, Sheldon J},
  journal={Bulletin of the American Meteorological Society},
  volume={81},
  number={6},
  pages={1241--1260},
  year={2000},
  publisher={American Meteorological Society}
}

@article{kossin2007globally,
  title={A globally consistent reanalysis of hurricane variability and trends},
  author={Kossin, Jim P and Knapp, Kenneth R and Vimont, Daniel J and Murnane, Richard J and Harper, Bruce A},
  journal={Geophysical Research Letters},
  volume={34},
  number={4},
  year={2007},
  publisher={Wiley Online Library}
}

@article{fetanat2013objective,
  title={Objective tropical cyclone intensity estimation using analogs of spatial features in satellite data},
  author={Fetanat, Gholamreza and Homaifar, Abdollah and Knapp, Kenneth R},
  journal={Weather and forecasting},
  volume={28},
  number={6},
  pages={1446--1459},
  year={2013}
}

@article{ritchie2014satellite,
  title={Satellite-derived tropical cyclone intensity in the North Pacific Ocean using the deviation-angle variance technique},
  author={Ritchie, Elizabeth A and Wood, Kimberly M and Rodr{\'\i}guez-Herrera, Oscar G and Pi{\~n}eros, Miguel F and Tyo, J Scott},
  journal={Weather and forecasting},
  volume={29},
  number={3},
  pages={505--516},
  year={2014},
  publisher={American Meteorological Society}
}

@article{liu2015satellite,
  title={A satellite-derived typhoon intensity index using a deviation angle technique},
  author={Liu, Chung-Chih and Liu, Chian-Yi and Lin, Tang-Huang and Chen, Liang-De},
  journal={International Journal of Remote Sensing},
  volume={36},
  number={4},
  pages={1216--1234},
  year={2015},
  publisher={Taylor \& Francis}
}

@inproceedings{chen2021real,
  title={Real-time tropical cyclone intensity estimation by handling temporally heterogeneous satellite data},
  author={Chen, Boyo and Chen, Buo-Fu and Chen, Yun-Nung},
  booktitle={Proceedings of the AAAI conference on artificial intelligence},
  volume={35},
  number={17},
  pages={14721--14728},
  year={2021}
}

@article{zhao2016multiple,
  title={A multiple linear regression model for tropical cyclone intensity estimation from satellite infrared images},
  author={Zhao, Yong and Zhao, Chaofang and Sun, Ruyao and Wang, Zhixiong},
  journal={Atmosphere},
  volume={7},
  number={3},
  pages={40},
  year={2016},
  publisher={MDPI}
}

@inproceedings{miller2017using,
  title={Using deep learning for tropical cyclone intensity estimation},
  author={Miller, Jeffrey and Maskey, Manil and Berendes, Todd},
  booktitle={AGU fall meeting abstracts},
  volume={2017},
  pages={IN11E--05},
  year={2017}
}

@article{pradhan2017tropical,
  title={Tropical cyclone intensity estimation using a deep convolutional neural network},
  author={Pradhan, Ritesh and Aygun, Ramazan S and Maskey, Manil and Ramachandran, Rahul and Cecil, Daniel J},
  journal={IEEE Transactions on Image Processing},
  volume={27},
  number={2},
  pages={692--702},
  year={2017},
  publisher={IEEE}
}

@InProceedings{tiny_vit,
  title={TinyViT: Fast Pretraining Distillation for Small Vision Transformers},
  author={Wu, Kan and Zhang, Jinnian and Peng, Houwen and Liu, Mengchen and Xiao, Bin and Fu, Jianlong and Yuan, Lu},
  booktitle={European conference on computer vision (ECCV)},
  year={2022}
}

@inproceedings{he2016deep,
  title={Deep residual learning for image recognition},
  author={He, Kaiming and Zhang, Xiangyu and Ren, Shaoqing and Sun, Jian},
  booktitle={Proceedings of the IEEE conference on computer vision and pattern recognition},
  pages={770--778},
  year={2016}
}

@article{zhai2014dependence,
  title={Dependence of US hurricane economic loss on maximum wind speed and storm size},
  author={Zhai, Alice R and Jiang, Jonathan H},
  journal={Environmental Research Letters},
  volume={9},
  number={6},
  pages={064019},
  year={2014},
  publisher={IOP Publishing}
}

@article{sun2017impact,
  title={Impact of ocean warming on tropical cyclone size and its destructiveness},
  author={Sun, Yuan and Zhong, Zhong and Li, Tim and Yi, Lan and Hu, Yijia and Wan, Hongchao and Chen, Haishan and Liao, Qianfeng and Ma, Chen and Li, Qihua},
  journal={Scientific reports},
  volume={7},
  number={1},
  pages={8154},
  year={2017},
  publisher={Nature Publishing Group UK London}
}

@article{lecun2015deep,
  title={Deep learning},
  author={LeCun, Yann and Bengio, Yoshua and Hinton, Geoffrey},
  journal={nature},
  volume={521},
  number={7553},
  pages={436--444},
  year={2015},
  publisher={Nature Publishing Group UK London}
}

@article{mendelsohn2012impact,
  title={The impact of climate change on global tropical cyclone damage},
  author={Mendelsohn, Robert and Emanuel, Kerry and Chonabayashi, Shun and Bakkensen, Laura},
  journal={Nature climate change},
  volume={2},
  number={3},
  pages={205--209},
  year={2012},
  publisher={Nature Publishing Group UK London}
}

@article{rappaport2000loss,
  title={Loss of life in the United States associated with recent Atlantic tropical cyclones},
  author={Rappaport, Edward N},
  journal={Bulletin of the American Meteorological Society},
  volume={81},
  number={9},
  pages={2065--2074},
  year={2000},
  publisher={American Meteorological Society}
}

@article{velden2006dvorak,
  title={The Dvorak tropical cyclone intensity estimation technique: A satellite-based method that has endured for over 30 years},
  author={Velden, Christopher and Harper, Bruce and Wells, Frank and Beven, John L and Zehr, Ray and Olander, Timothy and Mayfield, Max and Guard, Charles “CHIP” and Lander, Mark and Edson, Roger and others},
  journal={Bulletin of the American Meteorological Society},
  volume={87},
  number={9},
  pages={1195--1210},
  year={2006},
  publisher={American Meteorological Society}
}

@inproceedings{liu2021swin,
  title={Swin transformer: Hierarchical vision transformer using shifted windows},
  author={Liu, Ze and Lin, Yutong and Cao, Yue and Hu, Han and Wei, Yixuan and Zhang, Zheng and Lin, Stephen and Guo, Baining},
  booktitle={Proceedings of the IEEE/CVF international conference on computer vision},
  pages={10012--10022},
  year={2021}
}

@inproceedings{graham2021levit,
  title={Levit: a vision transformer in convnet's clothing for faster inference},
  author={Graham, Benjamin and El-Nouby, Alaaeldin and Touvron, Hugo and Stock, Pierre and Joulin, Armand and J{\'e}gou, Herv{\'e} and Douze, Matthijs},
  booktitle={Proceedings of the IEEE/CVF international conference on computer vision},
  pages={12259--12269},
  year={2021}
}

@inproceedings{howard2019searching,
  title={Searching for mobilenetv3},
  author={Howard, Andrew and Sandler, Mark and Chu, Grace and Chen, Liang-Chieh and Chen, Bo and Tan, Mingxing and Wang, Weijun and Zhu, Yukun and Pang, Ruoming and Vasudevan, Vijay and others},
  booktitle={Proceedings of the IEEE/CVF international conference on computer vision},
  pages={1314--1324},
  year={2019}
}

@article{xiao2021early,
  title={Early convolutions help transformers see better},
  author={Xiao, Tete and Singh, Mannat and Mintun, Eric and Darrell, Trevor and Doll{\'a}r, Piotr and Girshick, Ross},
  journal={Advances in neural information processing systems},
  volume={34},
  pages={30392--30400},
  year={2021}
}

@inproceedings{wu2021rethinking,
  title={Rethinking and improving relative position encoding for vision transformer},
  author={Wu, Kan and Peng, Houwen and Chen, Minghao and Fu, Jianlong and Chao, Hongyang},
  booktitle={Proceedings of the IEEE/CVF International Conference on Computer Vision},
  pages={10033--10041},
  year={2021}
}

@article{hendrycks2016gaussian,
  title={Gaussian error linear units (gelus)},
  author={Hendrycks, Dan and Gimpel, Kevin},
  journal={arXiv preprint arXiv:1606.08415},
  year={2016}
}

@inproceedings{ioffe2015batch,
  title={Batch normalization: Accelerating deep network training by reducing internal covariate shift},
  author={Ioffe, Sergey and Szegedy, Christian},
  booktitle={International conference on machine learning},
  pages={448--456},
  year={2015},
  organization={pmlr}
}

@article{ba2016layer,
  title={Layer normalization},
  author={Ba, Jimmy Lei and Kiros, Jamie Ryan and Hinton, Geoffrey E},
  journal={arXiv preprint arXiv:1607.06450},
  year={2016}
}

\newpage
\section*{Appendix}


\subsection*{Dataset}
TCIR gathers tropical cyclone data from satellite images across four channels: infrared (IR), water vapor (WV), passive microwave (PMW), and visible light (VIS). Each frame contains 201 $\times$ 201 data points, alongside corresponding wind speed information.

For our experiment setup, we employ typhoon images spanning the years 2003 to 2014 as the training dataset. Data from 2015 to 2016 are designated for validation, and the data from 2017 are reserved for testing. The training dataset comprises 40,348 frames corresponding to 730 typhoons, while the validation dataset encompasses 7,569 frames associated with 131 typhoons. The test set comprises 4,580 frames covering 94 typhoons.


\subsection*{Training strategy}
Before feeding an image into the model, we apply preprocessing steps, which include resizing the image to 224x224 pixels, performing a random rotation within the range of [0, ${20^\circ}$], and randomly applying horizontal or vertical flips with a 50\% probability.

To address potential overfitting, the Tint model's backbone is pretrained on ImageNet. We utilize the Mean Squared Error (MSE) as the loss function for network training. Our training setup employs a batch size of 32 and an initial learning rate of 0.00001. The learning rate undergoes a 10-fold decay at the 50th and 75th epochs, respectively. The training process continues for a total of 100 epochs.

\subsection*{Qualitative results}
\begin{figure}[h]
\centering
\includegraphics[width=135mm]{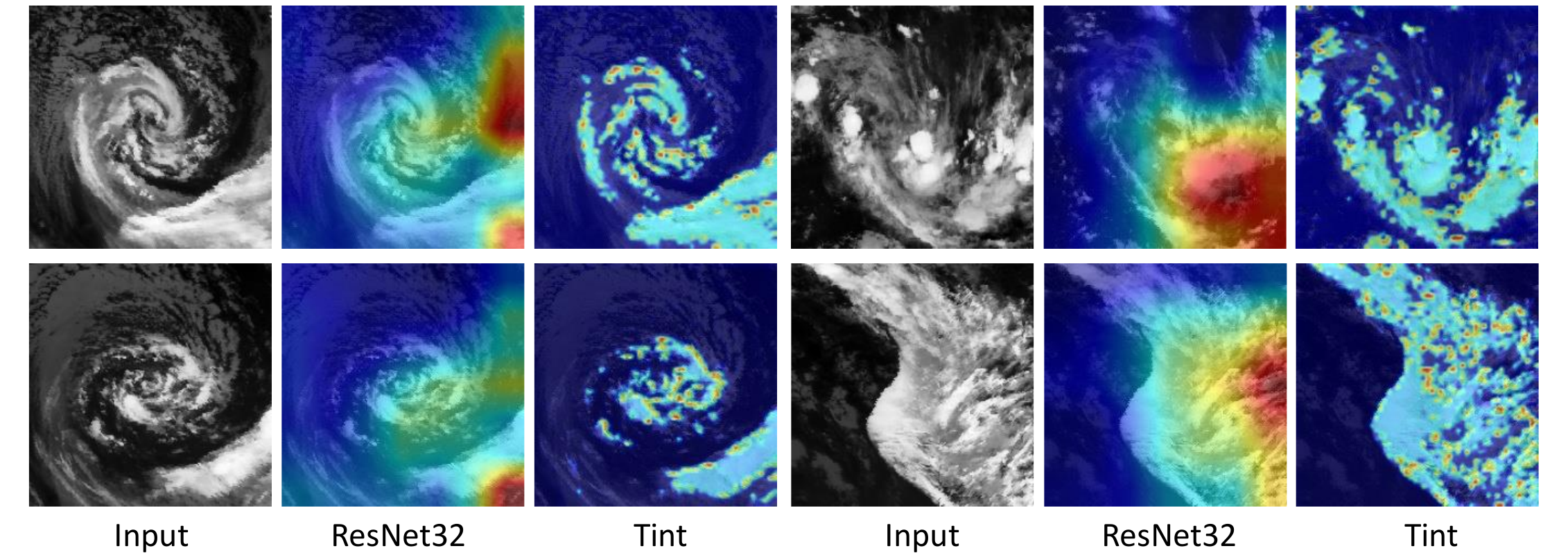}
\caption{The Grad-CAM visualization of ResNet32 and Tint.}
\label{fig2:env}
\end{figure}

\subsection*{Competitors}
Among the conventional meteorological models, we include:
Linear Regression-based Advanced Dvorak Technique (ADT) \cite{velden1998development}.
Advanced Microwave Sounding Unit (AMSU) \cite{kidder2000satellite}, which relies on near-earth orbit satellites and operates only during satellite passages through typhoons.
SATellite CONsensu (SATCON) \cite{velden2014update}, which is a heuristic combination of ADT and AMSU, widely used in forecasting practice, in particular relying on low-earth orbit satellite observations and expert inputs.
Kossin Technique (KT) \cite{kossin2007globally}, which improves predictive performance by constructing a global record of typhoon/hurricane intensity from existing data.
Feature Analogs in Satellite Imagery (FASI) \cite{fetanat2013objective}, which employs a $k$-nearest-neighbor algorithm to determine intensity based on the ten closest typhoons.
Improved DAVT \cite{ritchie2014satellite}, which predicts typhoon intensity through statistical analysis of the gradients of the IR brightness temperatures.
TI index \cite{liu2015satellite}, which leverages image edge processing techniques to examine meaningful discontinuity characteristics and calculate the brightness temperature gradients for typhoon intensity prediction.

We also consider the recently developed deep learning methods for our task:
Multiple Linear Regression Model (MLRM) \cite{zhao2016multiple}, which designs multiple features, including eyewall slope and brightness temperatures, and aggregates multiple regression models using different features for typhoon intensity prediction.
Transferring Deep Convolutional Neural Networks (TDCNN) \cite{miller2017using}, which introduces a simple Convolutional Neural Network (CNN).
Deep CNN \cite{pradhan2017tropical}, which proposes a deep neural network with regularization techniques.
CNN-TC \cite{chen2018rotation}, which presents a multi-modality deep network that takes both IR and PMW inputs meanwhile incorporating rotation-blending and sequence-smoothing techniques.
GAN-CNN \cite{chen2021real}, which employs Generative Adversarial Networks (GANs) to generate high-quality images, reducing the reliance on PMW data.
ResNet \cite{he2016deep}, which is a widely adopted CNN architecture with numerous successful applications in various domains.


\end{document}